\documentclass[11pt, a4paper, logo, twocolumn, copyright]{googledeepmind}

\pdftrailerid{redacted}

\makeatletter
\renewcommand\bibentry[1]{\nocite{#1}{\frenchspacing\@nameuse{BR@r@#1\@extra@b@citeb}}}
\makeatother

\usepackage{kantlipsum, lipsum}
\usepackage{dsfont}
\usepackage{gdm-colors}

\usepackage[authoryear, sort&compress, round]{natbib}

\usepackage{xspace}
\usepackage{multirow}

\DeclareMathOperator*{\argmin}{arg\,min}

\makeatletter
\DeclareRobustCommand\onedot{\futurelet\@let@token\@onedot}
\def\@onedot{\ifx\@let@token.\else.\null\fi\xspace}

\def\eg{\emph{e.g}\onedot}

 \def\vs{\emph{vs}\onedot}

\makeatother

\graphicspath{{figures/}}

\title{Video Creation by Demonstration}

\correspondingauthor{Ting Liu (\href{mailto:liuti@google.com}{liuti@google.com}) and Long Zhao (\href{mailto:longzh@google.com}{longzh@google.com}).}


\renewcommand{\today}{2024-12-09}

\author[1,2]{Yihong Sun}
\author[1]{Hao Zhou}
\author[1]{Liangzhe Yuan}
\author[1,2]{Jennifer J.\ Sun}
\author[1]{Yandong Li}
\author[1]{Xuhui Jia}
\author[1]{Hartwig Adam}
\author[2]{Bharath Hariharan}
\author[1]{Long Zhao}
\author[1]{Ting Liu}

\affil[1]{Google DeepMind}
\affil[2]{Cornell University}

  \newcommand{\ours}{$\delta$-Diffusion\xspace}
\newcommand{\vcbd}{Video Creation by Demonstration\xspace}
\makeatletter\renewcommand\paragraph{\@startsection{paragraph}{4}{\z@}
  {0.4em \@plus.0ex \@minus.2ex}{-.5em}{\normalfont\normalsize\bfseries}}\makeatother

\newcommand\projectpage{https://delta-diffusion.github.io}

\begin{abstract}
We explore a novel video creation experience, namely \emph{\vcbd}. Given a demonstration video and a context image from a different scene, we generate a physically plausible video that continues naturally from the context image and carries out the action concepts from the demonstration.
To enable this capability, we present \ours, a self-supervised training approach that learns from unlabeled videos by conditional future frame prediction.
Unlike most existing video generation controls that are based on explicit signals, we adopts the form of implicit latent control for maximal flexibility and expressiveness required by general videos. By leveraging a video foundation model with an appearance bottleneck design on top, we extract action latents from demonstration videos for conditioning the generation process with minimal appearance leakage.
Empirically, \ours outperforms related baselines in terms of both human preference and large-scale machine evaluations, and demonstrates potentials towards interactive world simulation.
Sampled video generation results are available at \url{\projectpage}.

\end{abstract} 
\begin{document}

\twocolumn[{\renewcommand\twocolumn[1][]{#1}\maketitle
\begin{center}
    \centering
    \captionsetup{type=figure}
\includegraphics[width=1\textwidth]{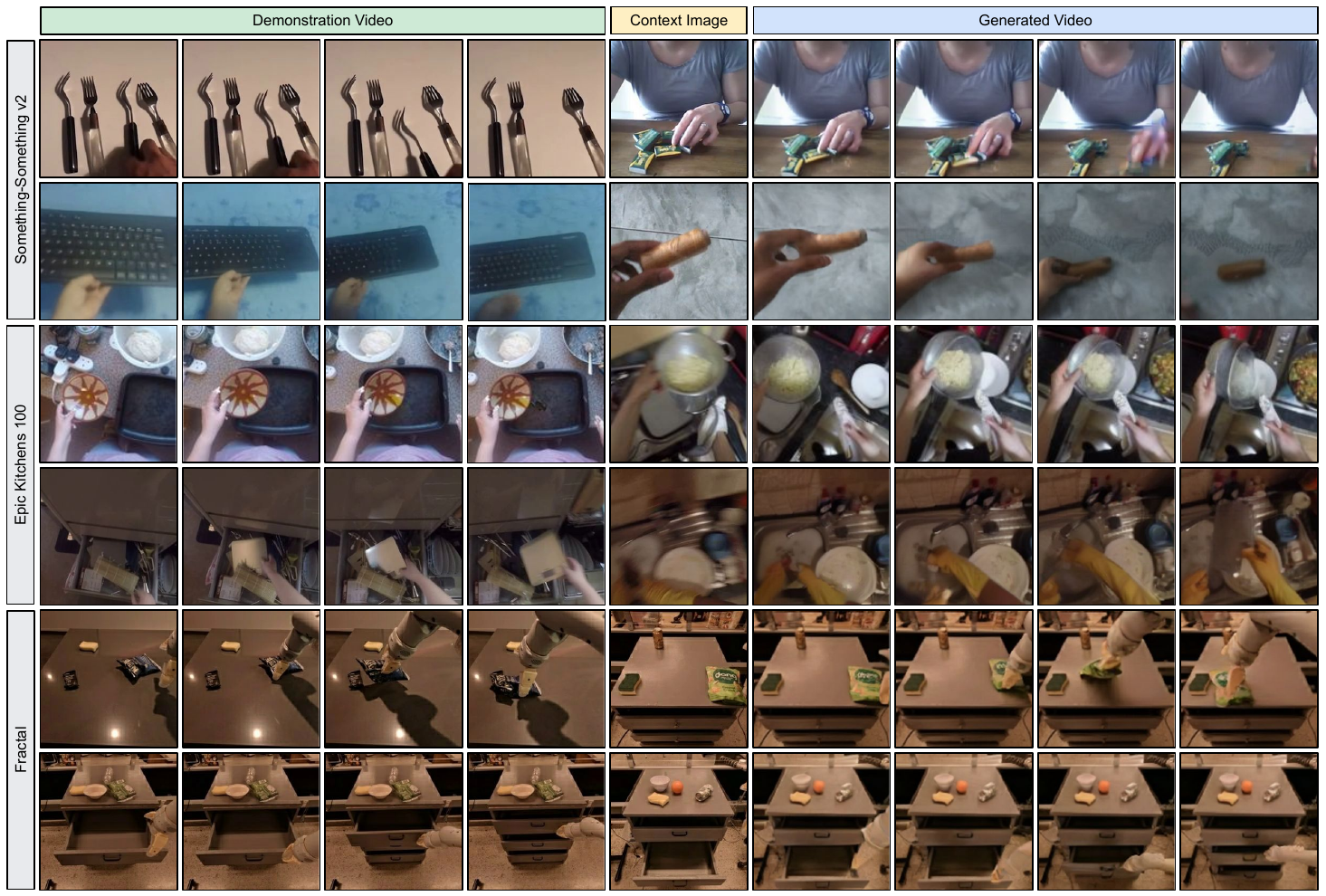}
\captionof{figure}{\textbf{\vcbd.} Given a demonstration video, our proposed \ours generates a video that naturally continues from a context image and carries out the same action concepts.
}
    \label{fig:teaser}
\end{center}}]
\newpage
\clearpage

\section{Introduction}
\label{sec:intro}

When given a visual demonstration, humans can naturally imagine what it would look like if these actions were to take place in a different environment.
This leads to a natural question, can we teach machines to simulate a realistic visual world with fine-grained action controls, all from a provided demonstration and a specified environment?
As an initial step towards answering this question, we propose \emph{\vcbd}, a video creation experience that empowers users to generate videos by providing a demonstration video that showcases desired action concepts and an initial scene context image to carry out the action concepts from.
As shown in Figure~\ref{fig:teaser}, our system generates a new video that integrates the demonstrated action into the provided context, ensuring both temporal continuity and physical plausibility.

With the recent advances in diffusion models~\citep{ho2020denoising,song2020score}, video generation~\citep{videoworldsimulators2024,polyak2024movie,walt} emerges as a frontier in the goal of building interactive world simulators~\citep{videoworldsimulators2024,bruce2024genie,valevski2024diffusion,alonso2024diffusion,meng2024towards}. 
Compared to our proposed \vcbd, most existing controllable video generation approaches are typically restricted to synthesizing dynamics through explicit control signals. 
These control signals, either abstract by nature (\eg, text prompts~\citep{walt}, moving keypoints~\citep{wu2025draganything}) or difficult to acquire (\eg, dense depth~\citep{chen2023control} or segmentation maps~\citep{han2022show}), limit user expressiveness or experience in interactive video creation.

Unlike previous approaches that focused on gaming~\citep{bruce2024genie,valevski2024diffusion} or domain-specific content like dancing or talking-head videos~\citep{song2018talking,siarohin2021motion}, we target general videos in this work, which presents significant challenges to video generation because actions in general videos are naturally contextualized and highly complex.
Consider Figure~\ref{fig:teaser}, the same action concept can appear drastically different depending on the subject performing the action, the object being acted upon, and the surrounding environment involved. 
This misalignment, along with inherent complexities in videos such as camera view changes and motion blurs, poses significant challenges for transferring action concepts between contexts.
On one hand, this fundamentally differentiates our work from conventional action transfer and retargeting~\citep{siarohin2019first,siarohin2021motion,ren2021flow}, which typically assume strict alignment between both actions and contexts of reference and target scenes.
On the other hand, this misalignment prevents us from mining paired training data as in previous methods~\citep{siarohin2021motion,ren2021flow}, making it challenging for model training.

To address these challenges, we propose three key designs. First, we enable deep understanding of complex actions and scenes in a demonstration video by leveraging the advancement of state-of-the-art video foundation models~\citep{wang2022internvideo,zhao2024videoprism,wang2024internvideo2,bardes2024revisiting}. Second, we adopt the form of implicit action latents to condition the generation process in place of explicit control signals to maximize the flexibility and expressiveness required by general videos. Third, we propose a self-supervised training paradigm for learning \vcbd. From a single video, we sample a context frame and its following clip as the demonstration to guarantee the context-action alignment, and task the model to generate the same clip. With these designs, we unlock new capabilities for video generation and manipulation beyond explicit signals, allowing for more nuanced and flexible control of video content.

In this paper, we introduce \ours, a novel two-stage training approach.
In the first stage, we extract spatiotemporal semantic representations of a demonstration video using a pre-trained video foundation model~\citep{zhao2024videoprism}. Directly conditioned on such representations during generation model training would lead to degenerate solutions where the context image is ignored. To tackle this issue, we learn an appearance bottleneck module on top of the video representations, by which we are able to extract action latents with minimal appearance/contextual information.
In the second stage, we train a diffusion model to predict future frames given the action concepts.
By conditioning the generation model on both the extracted control latents and a context image, we generate videos with realistic motion that seamlessly integrates with the specified action and context.
In contrast with supervised methods that require paired training data, \ours can potentially leverage large amounts of unlabeled video data for scaling.

We demonstrate the effectiveness of our approach through extensive experiments on diverse video datasets.
Our method is evaluated in terms of visual quality, action transferability, and context consistency through both machine and human evaluations. 
Notably, \ours is capable of generating high-fidelity videos spanning a wide range of action concepts, from everyday activities and ego-centric perspectives to complex robotic movements. We also show that creating videos by visual demonstration yields better controllability and concept transferability compared to text control. Furthermore, we are able to use different demonstration videos simply concatenated together to drive the generation of a coherent sequence, indicating the potential of leveraging \ours as an alternative to generative interactive environment~\citep{bruce2024genie}.

In summary, we make the following contributions. 
(i) We introduce \vcbd, a new creation experience for controllable video generation, which enables directly using videos as driving control signals for transferring action concepts.
(ii) To the best of our knowledge, we are the first to leverage out-of-the-box video foundation models for latent control of video generation. 
(iii) We propose a novel self-supervised approach for model training, which achieves compelling controllable video generation results.
Although some limitations remain (\eg, the result might not fully follow physical laws under complex scenes), we hope the proposed paradigm for controllable video generation will open new doors to interactive world simulation.
 \section{Related Works}
\label{sec:related}

\subsection{Video Generation}

Recent years have witnessed significant progress in video generation~\citep{blattmann2023align,harvey2022flexible,ho2022imagen,ho2022video,singer2022make,videoworldsimulators2024,polyak2024movie,walt}, with a range of methods for controlling the generated videos, including text-to-video~\citep{ramesh2021zero, ho2022imagen, singer2022make}, image-to-video~\citep{zhao2018learning, singer2022make}, image+text-to-video~\citep{walt, xiang2024pandora, wang2024motionctrl}, and image+video-to-video (I+V2V)~\citep{zhao2023motiondirector}.
Specifically, I+V2V methods include animating a still input image to create a video, conditioned on another video itself or some signals derived from another video. 
Our work, \vcbd, falls into the I+V2V category, where we aim to generate videos that continues from the context image while integrating the action concepts from the demonstration video.

Existing I+V2V techniques typically extract either explicit control signals (\eg, keypoints~\citep{chang2023magicpose}) or implicit control signals (\eg, learned embeddings~\citep{bruce2024genie}) from the condition video to influence the generated motion. 
These signals are then used alongside the initial frame to achieve I+V2V.
A common method that existing works have used to tackle I+V2V is to extract explicit control signals from the demonstration video, and use those to animate the input context frame. 
These extracted signals include text prompts~\citep{wang2024videocomposer, yang2023learning, xiang2024pandora, hu2023gaia}, depth maps and edge maps~\citep{wang2024videocomposer, chen2023control}, box or point tracks~\citep{li2024imageConductor, chen2023motion, wu2025draganything, wang2024motionctrl,wang2024boximator}, human keypoints~\citep{hu2024animate, park2024spectral}, or segmentation masks~\citep{xiao2024video, huang2022layered, davtyan2022controllable}. 
This also includes works that learn to leverage multiple types of control signals (\eg, sketch, depth, style), such as~\citep{wang2024videocomposer, chen2023control}.

Compared to extracting explicit signals from the control video, extracting implicit signals is comparatively less well-explored. 
One example is MotionDirector~\citep{zhao2023motiondirector}, which fine-tunes a video diffusion model for each demonstration video at test-time to generate the reference motion pattern that contains textures consistent with a given image. 
In comparison, our \ours aims to generate videos that naturally continue from the given context image, with no optimization during inference.
Another example is Genie~\citep{bruce2024genie}, which learns re-usable latent action codebook from video games. 
In addition to interactive generation where a sequence of discrete latent codes are inputted by users, these action codes can also be extracted from demonstration video to control generation.
In comparison, \ours extracts and utilizes action concepts that are more abstract than consecutive frame changes, while applying to the real-world visual domain.

\subsection{Modeling Motion and Action}
A key challenge in our framework is to effectively capture the action concept from the condition video, and transfer it to a different initial context.

One line of prior work has studied a task named action transfer or action re-targeting~\citep{song2018talking,siarohin2019first,siarohin2021motion,ren2021flow}, which often involves decomposing videos into motion and content representation, then transferring learned motion representations from one video into the content of another. While these methods have shown promising results in transferring motion patterns, they often operate under the assumption of a certain degree of alignment between the source and target videos (\eg, humans facing camera at the same distance). This alignment allows the low-level motion information to effectively represent higher-level actions.

However, this assumption does not hold in our setting of \vcbd, where the condition video and the target context can be  misaligned. In such cases, directly transferring low-level motion patterns may not accurately convey the intended action concept (\eg, video of a robot arm closing the top open drawer, to a robot arm closing the bottom open drawer).

\subsection{Video Foundation Models}

The emergence of powerful video foundation models, such as VideoPrism~\citep{zhao2024videoprism}, InternVideo~\citep{wang2022internvideo,wang2024internvideo2}, and V-JEPA~\citep{bardes2024revisiting}, has led to new directions for video understanding. These models are trained on massive datasets, learning to capture both low-level visual features and high-level semantic concepts. Unlike traditional video models that often focus on specific tasks like action recognition or object tracking, video foundation models are designed to be robust and adaptable, enabling them to be used for a wide range of downstream tasks~\citep{yuan2023videoglue}.

One of the key strengths of video foundation models is their ability to extract rich representations that capture both action and context information. This is demonstrated through their strong performance on various video understanding tasks, even when the foundation model is frozen. 
For example, VideoPrism~\citep{zhao2024videoprism} has shown state-of-the-art results on various tasks such as video captioning, question answering, and action localization using frozen features.
To the best of our knowledge, we are the first to study leveraging video foundation models for controlling I+V2V. \newcommand{\mV}{\mathcal{V}}
\newcommand{\mA}{\mathcal{A}}
\newcommand{\mB}{\mathcal{B}}
\newcommand{\mD}{\mathcal{D}}
\newcommand{\mP}{\mathcal{P}}
\newcommand{\mF}{\mathcal{F}}
\newcommand{\mG}{\mathcal{G}}
\newcommand{\mL}{\mathcal{L}}
\newcommand{\R}{\mathbb{R}}
\newcommand{\dv}{\delta(v)}

\begin{figure*}[t]
    \centering
    \includegraphics[width=\linewidth]{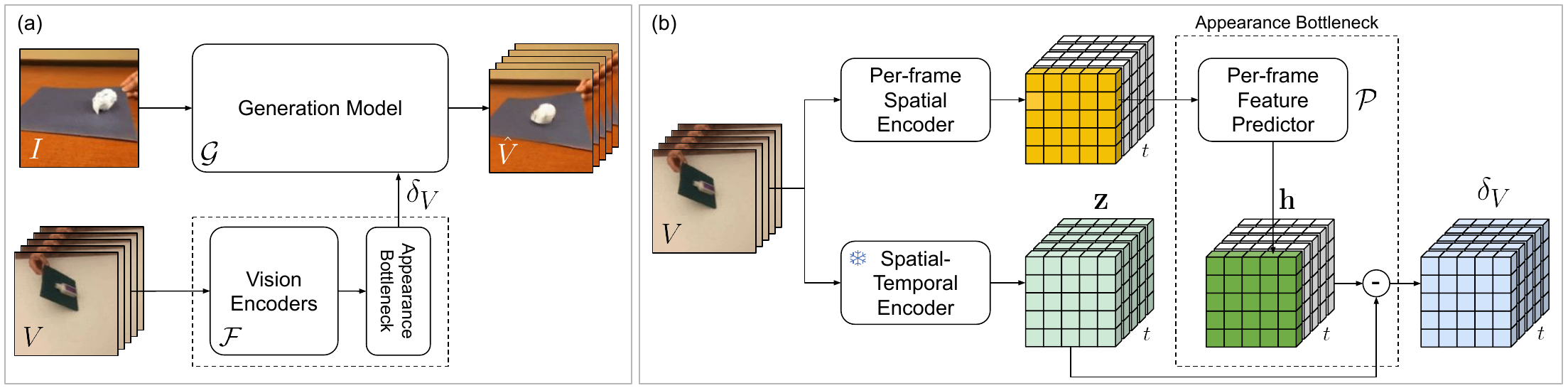}
    \caption{(a) \textbf{Overview of \ours.} The context frame $I$ is provided to the generation model $\mG$ along with the action latents $\delta_V$ extracted from the demonstration video $V$.
(b) \textbf{Extracting action latents.} A spatial-temporal vision encoder is applied to extract temporally-aggregated spatiotemopral representations $\mathbf{z}$ from an input video $V$, with $t$ denoting the temporal dimension.
    In parallel, a spatial vision encoder extracts per-frame representations from $V$, which is aligned to $\mathbf{z}$ by feature predictor $\mP$ as $\mathbf{h}$.
The appearance bottleneck then computes the action latents $\delta_V$ by subtracting the aligned spatial representations $\mathbf{h}$ from the spatiotemporal representations.
}
\label{fig:pipe_line_ab}
\end{figure*}

\section{Methodology}
\label{sec:method}

\subsection{Task Formulation}
\label{sec:task}

For the proposed task of \vcbd, the input is a context image $I$ providing contextual information and a demonstration video $V$ providing the control signal for generation. 
The goal is to generate a video $\hat{V}$ that naturally continues from the context image $I$ and carries out the action concepts in a similar manner as those found in the demonstration video $V$ (Figure~\ref{fig:teaser}).
We only consider this task valid when the action concepts are compatible with the input context.
For instance, we do not target simulating a cutting action in a context image without any hands in it.

\subsection{Method Overview}
\label{sec:overview}
Figure~\ref{fig:pipe_line_ab}(a) shows the overview of our proposed model \ours.
The generation model $\mG$ takes a context image $I$ and control latents extracted from a demonstration video $V$ as inputs, and outputs a desired video $\hat{V}$.
We apply vision encoders $\mF$ to compute spatiotemporal semantic representations of $V$, and one of our key designs is an appearance bottleneck applied on top to extract action-rich latents with minimal preservation of appearance information to condition the generation process (Section~\ref{sec:delta}). Conditioned on the bottlenecked control latents, \ours is trained in a self-supervised manner, where both the context image $I$ and the demonstration video $V$ are sampled from the same video, with $I$ being a starting frame followed by $V$, and the generation model $\mG$ is tasked to reconstruct $V$ as the target $\hat{V}$ (Section~\ref{sec:generation}).

\subsection{Extracting Action Latents}
\label{sec:delta}

Figure~\ref{fig:pipe_line_ab}(b) illustrates our design for extracting action control latents.
Given a demonstration video $V$, we apply a pre-trained video encoder to extract its temporally-aggregated semantic representations $\mathbf{z}\in \R^{T \times N \times D}$, where $T$, $N$, and $D$ represent the temporal, spatial, and feature dimension, respectively.
Such representations $\mathbf{z}$ typically is entangled with both appearance and action information, from which we propose to learn an appearance bottleneck to extract the desired action-rich control latents $\delta_V$.
The appearance bottleneck module consists of a feature predictor $\mP$ and a removal operator.
The feature predictor $\mP$ takes per-frame representations extracted from each frame of $V$ using a spatial encoder, and computes the best-effort approximation of the temporally-aggregated representations $\mathbf{z}$ independently for each frame.
The removal operator then subtracts the output of $\mP$ from the $\mathbf{z}$ to obtain the action-rich control latents $\delta_V$.

Our appearance bottleneck is designed following two principles: (i) the feature predictor $\mP$ extracts appearance representations from a single frame; (ii) the extracted appearance representations are ``compatible'' with the temporally-aggregated representations $\mathbf{z}$.
Principle (i) guarantees that $\mP$ extracts minimal action information, which is collectively defined by multiple consecutive frames.
Principle (ii) makes the design of the removal operator easy.

Based on these two principles, we formulate $\mP$ as a per-frame estimator that minimizes the difference between per-frame feature $\mathbf{z}_t$ and $\mathbf{h}_t$ over a dataset $\mD$ as
\begin{align}
    \mP &= \argmin_{\mP} \sum_{V\in \mD} \sum_{t=0}^{T-1} ||\mathbf{z}_t - \mathbf{h}_t||_1, \label{eq:bottleneck}
\end{align}
where $\mathbf{z}_t \in\R^{N \times D}$ denotes the $t$-th per-frame slice of $\mathbf{z}$, and $\mathbf{h}_t$ denotes the feature of the $t$-th frame from the input video $V$ predicted by $\mP$.
Intuitively, Equation~(\ref{eq:bottleneck}) learns $\mP$ that tries to reconstruct temporally-aggregated information based on a individual frame with best effort.
Without having access to the inter-frame ``action'' information, $\mP$ only captures the contextual appearance information.
Through reconstruction, $\mathbf{h}_t$ is aligned to the $\mathbf{z}_t$ space, and the removal operator can be simply designed as a subtraction operation.
In practice, we first extract visual features using a spatial encoder from each frame.
$\mP$ takes these visual features instead of raw frames as input, which makes its learning easier in practice.

The action control latents $\delta_V$ are then computed as
\begin{align}
    \delta_V &= \mathbf{z} - [\mathbf{h}_0, \mathbf{h}_1, \dots, \mathbf{h}_{T-1}], \label{eq:delta}
\end{align}
where $[\cdot]$ is a concatenation operation.
In Equation~(\ref{eq:delta}), $\delta_V$, the difference between the temporally-aggregated and per-frame information, represents the collective ``temporal surprisal'' that models the action concepts from the demonstration video, retrievable only through a sequence of frames.

\subsection{Training \ours}
\label{sec:generation}

We train the generation model $\mG$ in a self-supervised manner.
During training, we sample $T+1$ frames $v_{0:T}$ from a video $V$ in the training set, where $v_t$ denotes the $t$-th frame of $V$. The first frame $v_0$ is used as the context image $I$, and the following frames $v_{1:T}$ are used as the reconstruction target $\hat{V}$. $v_{1:T}$ are also used as the demonstration video, and the control latents are also extracted using the vision encoders and appearance bottleneck (Section~\ref{sec:delta}). In practice, we apply additional action-preserving random augmentations on $v_{1:T}$ into $V'$ before extracting the action latents $\delta_{V'}$ to further reduce appearance leakage. The augmentations, such as random spatial cropping, result in misalignment and disparity between the demonstration video and the target output during training, making it harder for $\mG$ to copy the appearance information directly from the demonstration video. $\mG$ is trained to predict $v_{1:T}$ from $v_0$ and $\delta_{V'}$ by minimizing the loss $\mL$
\begin{align}
    \mG &= \argmin_{\mG} \sum_{u\in \mD} \mathcal{L} \left( \mG(v_0, \delta_{V'}), v_{1:T} \right). \label{eq:diffusion}
\end{align}

The recent success of diffusion models for generating high quality videos~\citep{videoworldsimulators2024,polyak2024movie,walt} motivates us to adopt latent diffusion models (LDMs) as $\mG$. We first employ a tokenizer to compress both $v_0$ and $v_{1:T}$ into a low dimensional latent space. The LDM then takes the corrupted latents of $v_{1:T}$ together with other conditions as inputs and is trained with a  denoising loss function $\mL$. Following the practice in \cite{walt,salimans2022progressive}, we use the velocity as the prediction target of $\mL$. During inference, $\mG$ generates videos by iteratively denoising samples drawn from a noise distribution.

\subsection{Implementation Details}
\label{sec:implementation}

\paragraph{Vision encoders.} 
We adopt the video foundation model VideoPrism~\citep{zhao2024videoprism} (Base, $0.1$B parameters) as the vision encoders $\mF$.
VideoPrism adopts the factorized encoder architecture design from ViViT~\citep{arnab2021vivit}, which first computes the spatial representations per frame and then temporally aggregates each spatial location to output the final spatiotemporal representations.
This factorization in space and time allows the appearance bottleneck to directly take the per-frame spatial representations in VideoPrism as input, which reduces both training and inference cost.
VideoPrism takes $16$ $288\times 288$ frames as input and outputs a $16\times 16\times 16\times 768$ spatiotemporal representation tensor.

\paragraph{Appearance bottleneck feature predictor.}
The per-frame feature predictor $\mP$ is constructed as a stack of $4$ Transformer encoder blocks from ViViT~\citep{arnab2021vivit} with $1024$ hidden dimensions and $8$ heads for multi-head attention.
$\mP$ takes the intermediate VideoPrism spatial encoder output for a particular frame $v_t$ as input, and predicts its associated frame representations $\mathbf{h}_t$.
During training, we adopt a reconstruction objective with L$1$ loss and trained for $30$k iterations with the Adam optimizer~\citep{kingma2014adam} with $10^{-4}$ base learning rate and $4.5 \times 10^{-2}$ weight decay.

\paragraph{Video generation model.} 
For the video generation model $\mG$, we adopt WALT~\citep{walt} (Large, $0.3$B parameters). $\mG$ is trained to generate videos of $T=16$ frames with $128\times 128$ spatial resolution.
The context image $I$ is natively passed into the architecture as the image for conditional generation.
The control latents $\delta_V$ from demonstration video $V$ is projected into a sequence of $2048$-dimensional latent vectors in place of the original text embeddings for conditioning.
We initialize $\mG$ with a pre-trained I+T2V checkpoint, and fine-tune on downstream datasets in $500$k iterations. During inference, we use a classifier-free guidance scale of $1.25$~\citep{ho2022classifier}.

 \section{Experiment}
\label{sec:experiment}

\subsection{Datasets}
\label{sec:data}

We conduct our experiments primarily on three datasets, namely Epic Kitchens 100~\citep{ek100}, Something-Something v2 (SSv2)~\citep{goyal2017something}, and Fractal~\citep{fractal}.
We choose these datasets because of their rich human-object interactions and state changes in the video clips, making them a good testbed to demonstrate the proposed action concept transferring.
Epic Kitchens 100 is an egocentric video dataset, mostly focuses on kitchen scene. 
It features $55$ hours of videos with ground-truth annotations of fine-grained actions.
Something-Something v2 is a collection of labeled video clips of humans performing pre-defined basic actions with everyday objects.
It contains $174$ labeled actions in total with both first-person-view and third-person-view video clips and rich contents depicting human-object interactions.
Lastly, Fractal is a real-world robotics manipulation dataset firstly introduced in~\citep{fractal}.
It contains around $130$k episodes over $700$ tasks.
Each task is annotated by a simple natural language, such as ``pick redbull can from middle drawer and place on counter''.

To train our method, we first train a shared per-frame feature predictor $\mP$ on a mixture of the training split of all the datasets mentioned above.
We additionally include Ego4D~\citep{ego4d} during this training stage due to its rich and diversified content.
Our video generation model $\mG$ is then trained on each of the three datasets individually, yielding one model for each dataset.

The proposed \vcbd task requires a pair of \textit{<context image, demonstration video>}, and trivially picking one image paired with a random video during inference might lead to incompatible action concepts.
This is because the action conducted in the demonstration video could be completely infeasible to execute in the context image.
To mitigate this issue, we reorganize the existing datasets and curate meaningful pairs for validation.
Specifically, for each dataset, we first pick the examples with top-$10$ most frequent labels.
Within each label category, we then randomly sample $1$k video pairs. 
In each video pair $(\mathrm{A}, \mathrm{B})$, we take video $\mathrm{A}$ as the demonstration video and the first frame from video $\mathrm{B}$ as the context image. 
The generated video is evaluated against video $\mathrm{B}$ as we assume the videos with the same labeled description contains transferable action concept.
We use this curated large scale evaluation dataset for machine evaluation.
Additionally, we manually select a small set of video pairs by carefully verifying the transferability of their action concepts. Please refer to appendix for details of this selection process.
These selected video pairs are used for human evaluation.

\begin{figure}[t]
    \centering
    \includegraphics[width=0.49\textwidth]{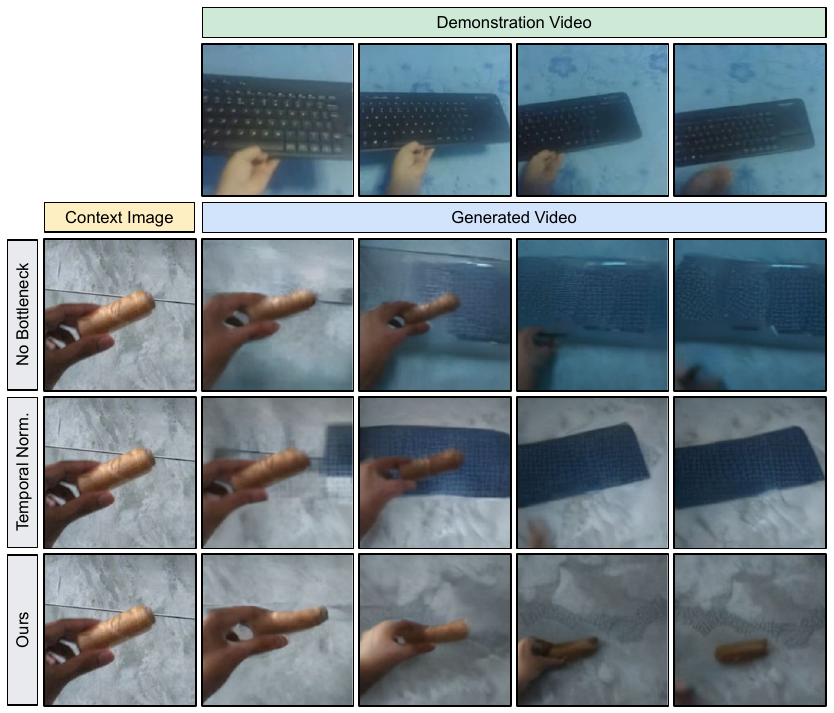}
    \caption{\textbf{Qualitative results for bottleneck ablation} on the Something-Something v2 dataset~\citep{goyal2017something}.
    Applying no or temporal normalization bottleneck suffers from appearance leakage, while generation based on our appearance bottleneck preserves the input context.
    }
    \label{fig:ablation}
\end{figure}

\subsection{Evaluation Setup}
\label{sec:experiment_eval}

We conduct human and machine evaluations to study the effectiveness of different methods.

\begin{table*}[tb]
    \caption{\textbf{Ablation study on \ours.} We ablate the proposed appearance bottleneck, and consider alternative variants that serve as competitive baselines.
    For the applied bottleneck, ``None'' indicates no bottleneck is placed while ``Temp.\ Norm.'' indicates temporal normalization applied to the spatiotemporal features. 
    We evaluate on Something-Something v2, Epic Kitchens 100, and Fractal for comparisons in terms of generation quality (FVD) and context-generation alignment via both embedding cosine similarity (ES) and retrieval Hit@k (\%).}
    \centering
    \resizebox{\linewidth}{!}{
    \begin{tabular}{@{}c  |  ccc  |  ccc  | ccc@{}}
        \toprule
         \multirow{2}{*}{Bottleneck} &\multicolumn{3}{c|}{SSv2} &\multicolumn{3}{c|}{Epic Kitchens 100} &\multicolumn{3}{c}{Fractal} \\
& FVD ($\downarrow$) & ES ($\uparrow$) & Hit@100/500/1k ($\uparrow$) & FVD ($\downarrow$) & ES ($\uparrow$) & Hit@100/500/1k  ($\uparrow$) & FVD ($\downarrow$) & ES ($\uparrow$) & Hit@100/500/1k ($\uparrow$)\\
        \midrule
         None              & 47.3  & 0.838 & 57.7 / 74.3 / 81.7 & 46.2  & 0.847 & 38.9 / 58.9 / 69.1 & \textbf{41.9}  & 0.902 & 58.0 / 75.7 / 82.8\\
         Temp.\ Norm.\           & 38.4  & 0.846 & 63.4 / 79.1 / 85.4 & \textbf{41.7}  & 0.850 & 41.7 / 61.3 / 70.7 & 42.1 & 0.906 & 62.2 / 78.2 / 85.0\\
Ours      & \textbf{38.0}  & \textbf{0.853} & \textbf{66.9} / \textbf{81.8} / \textbf{87.3} & 42.3 & \textbf{0.854} & \textbf{44.9} / \textbf{64.6} / \textbf{74.2} & 44.9 & \textbf{ 0.907} & \textbf{63.0} / \textbf{78.8} / \textbf{85.2}\\
        \bottomrule
    \end{tabular}}
    \label{tab:quant}
\end{table*}

\begin{table*}[tb]
    \caption{\textbf{Human evaluation preference rate on \ours.} We compare against two baseline methods in condition video generation that are relevant our task. We ask human raters to evaluate the performance in terms of visual quality (VQ), action transferability between demonstration and generated videos (AT), and context image consistency (CC).}
    \centering
    \resizebox{\linewidth}{!}{
    \begin{tabular}{@{}l c  |  ccc  |  ccc  | ccc@{}}
        \toprule
        \multirow{ 2}{*}{Method} & \multirow{ 2}{*}{Condition} &\multicolumn{3}{c|}{SSv2} &\multicolumn{3}{c|}{Epic Kitchens 100} &\multicolumn{3}{c}{Fractal} \\
& & VQ ($\uparrow$) & AT ($\uparrow$) & CC ($\uparrow$) & VQ ($\uparrow$) & AT ($\uparrow$) & CC ($\uparrow$) & VQ ($\uparrow$) & AT ($\uparrow$) & CC ($\uparrow$)\\
        \midrule
\vs WALT  & image+text & 0.74 & 0.77 & 0.70 & 0.70 & 0.83 & 0.65 & 0.82 & 0.80 & 0.74  \\
\vs MotionDirector  & image+text+video & 0.86 & 0.98 & 0.91 & 0.96 & 0.98 & 0.98 & 0.96 & 1.00 & 0.97  \\
\bottomrule
    \end{tabular}}
    \label{tab:human}
\end{table*} 
\paragraph{Machine evaluation.}  Pre-trained image/video understanding models are leveraged to quantify the quality of generated examples in machine evaluation.
It brings scalability advantages over the human evaluation which can be used to evaluate large cohort of generated examples.
In this study, we apply three quantitative machine evaluation metrics to evaluate the action concept transfer quality in different aspects.
We apply the commonly used Fr\'echet Video Distance (FVD)~\citep{unterthiner2018fvd} to measure the overall generation fidelity.
However, FVD only reflects the distribution shift between generated and reference videos and cannot measure whether the generated video faithfully follow action concept of reference video. 
To complement this, we apply I3D~\citep{carreira2017i3d} to compute features from the generated and reference videos and report their averaged embedding cosine similarity (ES).
Finally, we also construct a retrieval-based evaluation to measure the similarity between the generated and reference video in the context of all reference videos. We use generated video features and attempt to retrieve the corresponding reference video via embedding cosine similarity. 
The retrieval performance is evaluated in terms of hit-rate at $100$, $500$, and $1$k, which correspond to 1\%, 5\%, and  10\% of the retrieval set, respectively.

\paragraph{Human evaluation.} On the other hand, human evaluation captures the human preference on the task.  
In this studies, we hire five human raters who are trained to assess the overall visual quality (VQ), action transferability (AT) and content consistency (CC) of the generated videos.
The detailed description of each rubrics could be found in the appendix.
We curate $20$ video examples from each dataset, resulting in $60$ examples in total, as the human evaluation set.
In each turn, a reference image, a demonstration video, and two generated videos (one from the proposed method and the other from the method to compare) are provided to the rater.
The rater is then asked to choose the better video from the two generated ones based on each of the three rubrics mentioned above.
We then compute the preference rate of human rater favoring our method against the baseline under each dataset and rubric as the evaluation metric.

\begin{figure*}[t]
    \centering
    \includegraphics[width=\textwidth]{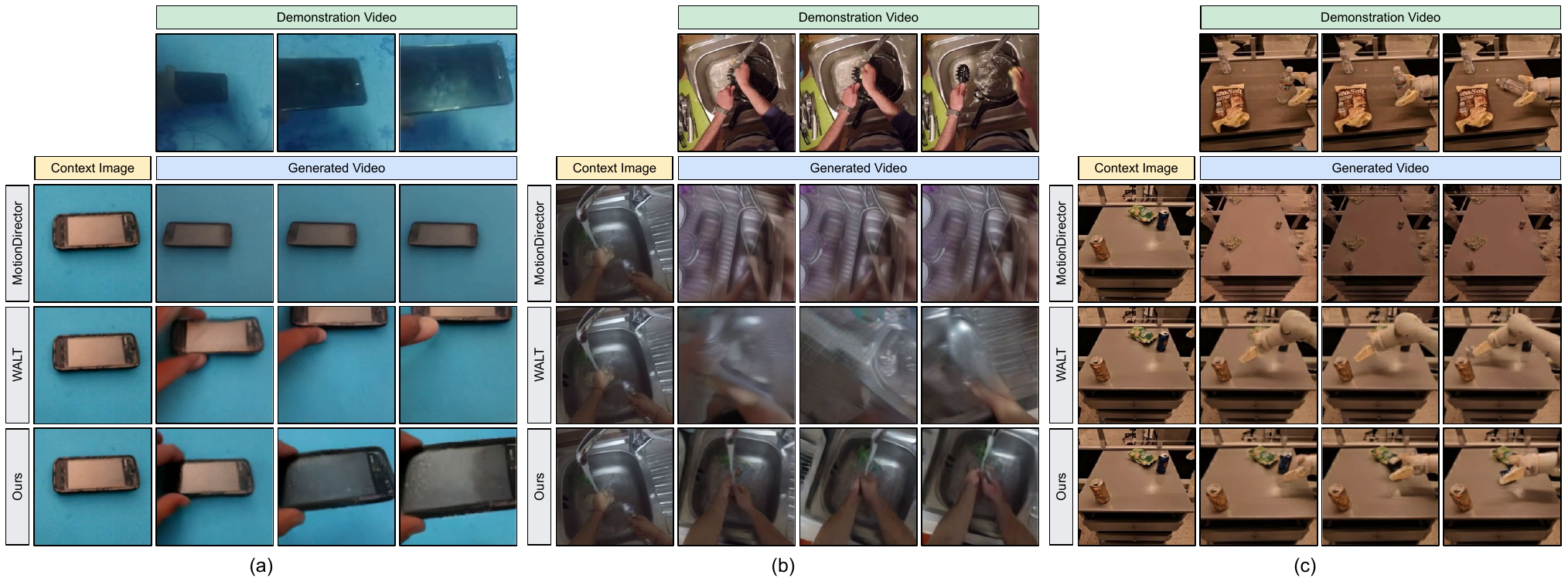}
    \vspace{-1.3em}
    \caption{\textbf{Qualitative comparisons} of \ours against MotionDirector~\citep{zhao2023motiondirector} and WALT~\citep{walt} on (a) Something-Something v2~\citep{goyal2017something}, (b) Epic Kitchens 100~\citep{ek100}, and (c) Fractal~\citep{fractal} datasets.}
    \label{fig:qual}
\end{figure*}

\begin{figure*}[t]
    \centering
    \includegraphics[width=\textwidth]{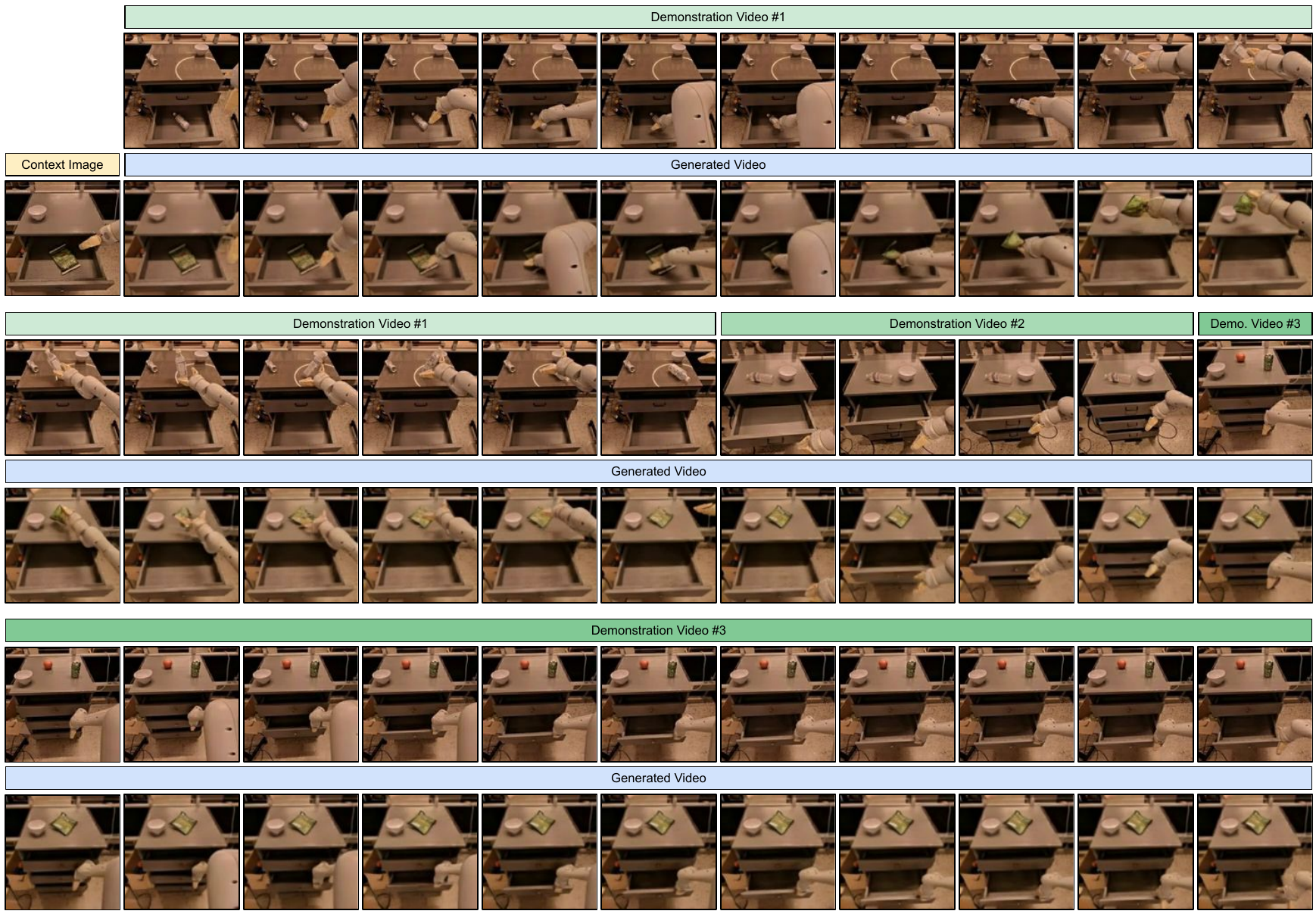}
    \vspace{-1.3em}
    \caption{\textbf{Auto-regressive generation} controlled via a concatenation of three different demonstration videos of varying lengths. The sequence of demonstrated action concepts (``picking something from a drawer and placing it on the table'', ``closing a drawer'', and ``opening a drawer'') are coherently transferred to the input context.}
    \label{fig:video_collage}
\end{figure*}

\subsection{Ablation Study}
In this section, we ablate the design choice of our training algorithm first.
We use large scale evaluation data as described in Section~\ref{sec:data} and machine metrics as described in Section~\ref{sec:experiment_eval} for the ablation study.
The results are presented in Table~\ref{tab:quant}.
We compare two variant bottleneck designs with our proposed appearance bottleneck.
``None'' stands for using features $\mathbf{z}$ as the condition signal and we do not apply any bottleneck training on the features.
In the temporal normalization setup, following \cite{xiao2024video}, the condition signal is obtained by firstly applying average pooling along the temporal dimension of $\mathbf{z}$ and then subtracting it from $\mathbf{z}$.

From the results, we observe that \ours performs better than two counter-parts. 
On instance-wise metrics, retrieval hit-rate (Hit@k) and embedding cosine similarity (ES), \ours consistently outperforms baselines with none or temporal normalization as the bottleneck.
This indicates an improved action concept transferability with our proposed appearance bottleneck design.
On the other hand, we do not observe consistent win among the three methods in terms of visual quality.
This shows that while increasing the action faithfulness and temporal consistency, the proposed method does not hurt the visual quality of the generated videos.
Figure~\ref{fig:ablation} further shows visual comparisons for different bottleneck designs. 
We notice that both without bottleneck and ``temporal norm.'' bottleneck suffer from appearance leakage; in contrast, the proposed appearance bottleneck successfully retain action concepts from the demonstration video while minimizing the appearance information.

\subsection{Main Results}

\paragraph{Comparison with baselines.} Other than video, text is also a natural format to describe actions.
In our study, we take WALT~\citep{walt} as the representative of video generation conditioned on image and text.
For a fair comparison, we fine-tune it on each dataset individually with ground truth captions.
We also compare the proposed method with MotionDirector~\citep{zhao2023motiondirector} which uses image, text, and video as conditional signals.
When generating videos, MotionDirector learns per-instance appearance and motion from context image and demonstration video via LoRA fine-tuning~\citep{hu2021lora}, and generate videos based on the given text prompt.
For both WALT and MotionDirector, we use the groundtruth caption label as the prompt during video generation.
Please refer to the appendix for more details of these two baseline.

We report the human evaluation results of ours against two baselines in Table~\ref{tab:human}. 
The numbers indicate the preference rate of human rater favoring our method against the baseline under each dataset and rubric.
We notice that \ours is clearly favored by the human raters across all the datasets and rubrics, which underlines its superior performance.

We compare the generated videos by our methods and other baselines in Figure~\ref{fig:qual}.
We note that MotionDirector would fail in many cases, possibly due to its difficulties in optimizing the LoRA parameters.
WALT, although preserves content consistency well, could carry out in-genuine action to the demonstration video. 
This could due to the limited expressiveness in text description on demonstration videos.
On our results, we show both faithful action execution and content preservation in the generated videos.

\paragraph{Auto-regressive generation.}
We further explore auto-regressively generating a coherent video that starts from a single context image, conditioned on a sequence of demonstration videos, as shown in Figure~\ref{fig:video_collage}.
At each step, the generation model uses the last frames from the previous step as the context.
Each conditioning video is selected from a different scene, but overall they demonstrate a complete rollout of a complex task.
We show that by sequentially applying the action latents from the demonstration videos, we are able to auto-regressively simulate the task execution in one consistent environment.

 \section{Conclusion}
\label{sec:conclusion}

In this work, we introduce \vcbd, a video creation experience that enables users to generate videos by providing an initial context frame and a demonstration video for action concept conditioning. 
The main challenge in this task is how to handle the misalignment of appearances and contexts between the demonstration video and the context image during the transferring of action concepts.
To address this, we propose to leverage the video foundation model to provide semantic representations for actions and contexts in videos and build an appearance bottleneck on top of them to extract latent control signals with rich action concept and minimum appearance information.
Then, a video generation model is learned to take the context image and the control signals as inputs and generate videos that continue from the context image and carry out the action concept in the similar manner as shown in the demonstration video.
Our extensive experiments on three separate datasets using both machine and human evaluations demonstrate the effectiveness of our proposed \ours.
One of the limitations is that \ours at its current form does not always strictly preserves physical realism under complex scenes, which needs further exploration and can be potentially alleviated by scaling up the generation model. 
\section*{Broader Impact}

\paragraph{Societal impact.}
Our proposed \vcbd improves the ease of interactive generation workflow, by offering a novel video creation experience that bridges the gap between abstract (less controllable) and detailed (hard to obtain) control signals.
In addition, \ours demonstrates capabilities of world modeling beyond the transfer of action concepts, by simulating possible effects of these actions on the entire scene (\eg, actions would cause additional interactions in the given context that is not captured by the demonstration video.)

\paragraph{Potential negative impact.}
Our work does not introduce any negative societal impacts beyond those commonly found in controlled video generation.
These may include reinforcement of existing dataset bias, misuse of generated content for creating misleading or inappropriate materials, and privacy concerns involving individuals without their explicit consent.
We highlight the importance of vigilance and responsible implementation to mitigate these impacts and guarantee ethical use.

\section*{Acknowledgements}
We sincerely thank Anqi Huang at UC Irvine, and Boqing Gong, Bohyung Han, David Hendon, Hexiang Hu, Jimin Pi, Luke Friedman, Luming Tang, Mikhail Sirotenko, Ming-Hsuan Yang, Mingda Zhang, Sanghyun Woo, Willis Ma, Yukun Zhu, Yuxiao Wang, and Ziyu Wan at Google DeepMind for their feedback, discussion, and support. We also thank Florian Schroff, Huisheng Wang, Caroline Pantofaru, and Tomas Izo for their leadership support for this project.

\newpage
\onecolumn

\bibliographystyle{abbrvnat}
\nobibliography*
\bibliography{main}

\begin{thebibliography}{57}
\providecommand{\natexlab}[1]{#1}
\providecommand{\url}[1]{\texttt{#1}}
\expandafter\ifx\csname urlstyle\endcsname\relax
  \providecommand{\doi}[1]{doi: #1}\else
  \providecommand{\doi}{doi: \begingroup \urlstyle{rm}\Url}\fi

\bibitem[Alonso et~al.(2024)Alonso, Jelley, Micheli, Kanervisto, Storkey,
  Pearce, and Fleuret]{alonso2024diffusion}
E.~Alonso, A.~Jelley, V.~Micheli, A.~Kanervisto, A.~Storkey, T.~Pearce, and
  F.~Fleuret.
\newblock Diffusion for world modeling: Visual details matter in {Atari}.
\newblock \emph{NeurIPS}, 2024.

\bibitem[Arnab et~al.(2021)Arnab, Dehghani, Heigold, Sun, Lu{\v{c}}i{\'c}, and
  Schmid]{arnab2021vivit}
A.~Arnab, M.~Dehghani, G.~Heigold, C.~Sun, M.~Lu{\v{c}}i{\'c}, and C.~Schmid.
\newblock {ViViT}: A video vision transformer.
\newblock In \emph{CVPR}, 2021.

\bibitem[Bardes et~al.(2024)Bardes, Garrido, Ponce, Chen, Rabbat, LeCun,
  Assran, and Ballas]{bardes2024revisiting}
A.~Bardes, Q.~Garrido, J.~Ponce, X.~Chen, M.~Rabbat, Y.~LeCun, M.~Assran, and
  N.~Ballas.
\newblock Revisiting feature prediction for learning visual representations
  from video.
\newblock \emph{TMLR}, 2024.

\bibitem[Blattmann et~al.(2023)Blattmann, Rombach, Ling, Dockhorn, Kim, Fidler,
  and Kreis]{blattmann2023align}
A.~Blattmann, R.~Rombach, H.~Ling, T.~Dockhorn, S.~W. Kim, S.~Fidler, and
  K.~Kreis.
\newblock Align your latents: High-resolution video synthesis with latent
  diffusion models.
\newblock In \emph{CVPR}, 2023.

\bibitem[Brohan et~al.(2022)Brohan, Brown, Carbajal, Chebotar, Dabis, Finn,
  Gopalakrishnan, Hausman, Herzog, Hsu, et~al.]{fractal}
A.~Brohan, N.~Brown, J.~Carbajal, Y.~Chebotar, J.~Dabis, C.~Finn,
  K.~Gopalakrishnan, K.~Hausman, A.~Herzog, J.~Hsu, et~al.
\newblock {RT-1}: Robotics transformer for real-world control at scale.
\newblock \emph{arXiv preprint arXiv:2212.06817}, 2022.

\bibitem[Brooks et~al.(2024)Brooks, Peebles, Holmes, DePue, Guo, Jing, Schnurr,
  Taylor, Luhman, Luhman, Ng, Wang, and Ramesh]{videoworldsimulators2024}
T.~Brooks, B.~Peebles, C.~Holmes, W.~DePue, Y.~Guo, L.~Jing, D.~Schnurr,
  J.~Taylor, T.~Luhman, E.~Luhman, C.~Ng, R.~Wang, and A.~Ramesh.
\newblock Video generation models as world simulators.
\newblock \emph{OpenAI Blog}, 2024.
\newblock URL
  \url{https://openai.com/research/video-generation-models-as-world-simulators}.

\bibitem[Bruce et~al.(2024)Bruce, Dennis, Edwards, Parker-Holder, Shi, Hughes,
  Lai, Mavalankar, Steigerwald, Apps, et~al.]{bruce2024genie}
J.~Bruce, M.~D. Dennis, A.~Edwards, J.~Parker-Holder, Y.~Shi, E.~Hughes,
  M.~Lai, A.~Mavalankar, R.~Steigerwald, C.~Apps, et~al.
\newblock Genie: Generative interactive environments.
\newblock In \emph{ICML}, 2024.

\bibitem[Carreira and Zisserman(2017)]{carreira2017i3d}
J.~Carreira and A.~Zisserman.
\newblock {Quo vadis, action recognition? A new model and the Kinetics
  dataset}.
\newblock In \emph{CVPR}, 2017.

\bibitem[Chang et~al.(2024)Chang, Shi, Gao, Xu, Fu, Song, Yan, Zhu, Yang, and
  Soleymani]{chang2023magicpose}
D.~Chang, Y.~Shi, Q.~Gao, H.~Xu, J.~Fu, G.~Song, Q.~Yan, Y.~Zhu, X.~Yang, and
  M.~Soleymani.
\newblock {MagicPose}: Realistic human poses and facial expressions retargeting
  with identity-aware diffusion.
\newblock In \emph{ICML}, 2024.

\bibitem[Chen et~al.(2023{\natexlab{a}})Chen, Zhang, and
  Hinton]{chen2022analog}
T.~Chen, R.~Zhang, and G.~Hinton.
\newblock {Analog Bits}: Generating discrete data using diffusion models with
  self-conditioning.
\newblock In \emph{ICLR}, 2023{\natexlab{a}}.

\bibitem[Chen et~al.(2023{\natexlab{b}})Chen, Lin, Tseng, Lin, and
  Yang]{chen2023motion}
T.-S. Chen, C.~H. Lin, H.-Y. Tseng, T.-Y. Lin, and M.-H. Yang.
\newblock Motion-conditioned diffusion model for controllable video synthesis.
\newblock \emph{arXiv preprint arXiv:2304.14404}, 2023{\natexlab{b}}.

\bibitem[Chen et~al.(2023{\natexlab{c}})Chen, Ji, Wu, Wu, Xie, Li, Xia, Xiao,
  and Lin]{chen2023control}
W.~Chen, Y.~Ji, J.~Wu, H.~Wu, P.~Xie, J.~Li, X.~Xia, X.~Xiao, and L.~Lin.
\newblock {Control-A-Video}: Controllable text-to-video generation with
  diffusion models.
\newblock \emph{arXiv preprint arXiv:2305.13840}, 2023{\natexlab{c}}.

\bibitem[Damen et~al.(2018)Damen, Doughty, Farinella, Fidler, Furnari, Kazakos,
  Moltisanti, Munro, Perrett, Price, et~al.]{ek100}
D.~Damen, H.~Doughty, G.~M. Farinella, S.~Fidler, A.~Furnari, E.~Kazakos,
  D.~Moltisanti, J.~Munro, T.~Perrett, W.~Price, et~al.
\newblock Scaling egocentric vision: The {EPIC-KITCHENS} dataset.
\newblock In \emph{ECCV}, 2018.

\bibitem[Davtyan and Favaro(2022)]{davtyan2022controllable}
A.~Davtyan and P.~Favaro.
\newblock Controllable video generation through global and local motion
  dynamics.
\newblock In \emph{ECCV}, 2022.

\bibitem[Goyal et~al.(2017)Goyal, Ebrahimi~Kahou, Michalski, Materzynska,
  Westphal, Kim, Haenel, Fruend, Yianilos, Mueller-Freitag,
  et~al.]{goyal2017something}
R.~Goyal, S.~Ebrahimi~Kahou, V.~Michalski, J.~Materzynska, S.~Westphal, H.~Kim,
  V.~Haenel, I.~Fruend, P.~Yianilos, M.~Mueller-Freitag, et~al.
\newblock The ``something something'' video database for learning and
  evaluating visual common sense.
\newblock In \emph{ICCV}, 2017.

\bibitem[Grauman et~al.(2022)Grauman, Westbury, Byrne, Chavis, Furnari,
  Girdhar, Hamburger, Jiang, Liu, Liu, et~al.]{ego4d}
K.~Grauman, A.~Westbury, E.~Byrne, Z.~Chavis, A.~Furnari, R.~Girdhar,
  J.~Hamburger, H.~Jiang, M.~Liu, X.~Liu, et~al.
\newblock {Ego4D}: Around the world in 3,000 hours of egocentric video.
\newblock In \emph{CVPR}, 2022.

\bibitem[Gupta et~al.(2024)Gupta, Yu, Sohn, Gu, Hahn, Li, Essa, Jiang, and
  Lezama]{walt}
A.~Gupta, L.~Yu, K.~Sohn, X.~Gu, M.~Hahn, F.-F. Li, I.~Essa, L.~Jiang, and
  J.~Lezama.
\newblock Photorealistic video generation with diffusion models.
\newblock In \emph{ECCV}, 2024.

\bibitem[Han et~al.(2022)Han, Ren, Lee, Barbieri, Olszewski, Minaee, Metaxas,
  and Tulyakov]{han2022show}
L.~Han, J.~Ren, H.-Y. Lee, F.~Barbieri, K.~Olszewski, S.~Minaee, D.~Metaxas,
  and S.~Tulyakov.
\newblock Show me what and tell me how: Video synthesis via multimodal
  conditioning.
\newblock In \emph{CVPR}, 2022.

\bibitem[Harvey et~al.(2022)Harvey, Naderiparizi, Masrani, Weilbach, and
  Wood]{harvey2022flexible}
W.~Harvey, S.~Naderiparizi, V.~Masrani, C.~Weilbach, and F.~Wood.
\newblock Flexible diffusion modeling of long videos.
\newblock In \emph{NeurIPS}, 2022.

\bibitem[Ho and Salimans(2022)]{ho2022classifier}
J.~Ho and T.~Salimans.
\newblock Classifier-free diffusion guidance.
\newblock \emph{arXiv preprint arXiv:2207.12598}, 2022.

\bibitem[Ho et~al.(2020)Ho, Jain, and Abbeel]{ho2020denoising}
J.~Ho, A.~Jain, and P.~Abbeel.
\newblock Denoising diffusion probabilistic models.
\newblock In \emph{NeurIPS}, 2020.

\bibitem[Ho et~al.(2022{\natexlab{a}})Ho, Chan, Saharia, Whang, Gao, Gritsenko,
  Kingma, Poole, Norouzi, Fleet, et~al.]{ho2022imagen}
J.~Ho, W.~Chan, C.~Saharia, J.~Whang, R.~Gao, A.~Gritsenko, D.~P. Kingma,
  B.~Poole, M.~Norouzi, D.~J. Fleet, et~al.
\newblock {Imagen Video}: High definition video generation with diffusion
  models.
\newblock \emph{arXiv preprint arXiv:2210.02303}, 2022{\natexlab{a}}.

\bibitem[Ho et~al.(2022{\natexlab{b}})Ho, Salimans, Gritsenko, Chan, Norouzi,
  and Fleet]{ho2022video}
J.~Ho, T.~Salimans, A.~Gritsenko, W.~Chan, M.~Norouzi, and D.~J. Fleet.
\newblock Video diffusion models.
\newblock In \emph{NeurIPS}, 2022{\natexlab{b}}.

\bibitem[Hu et~al.(2023)Hu, Russell, Yeo, Murez, Fedoseev, Kendall, Shotton,
  and Corrado]{hu2023gaia}
A.~Hu, L.~Russell, H.~Yeo, Z.~Murez, G.~Fedoseev, A.~Kendall, J.~Shotton, and
  G.~Corrado.
\newblock {GAIA-1}: A generative world model for autonomous driving.
\newblock \emph{arXiv preprint arXiv:2309.17080}, 2023.

\bibitem[Hu et~al.(2022)Hu, Shen, Wallis, Allen-Zhu, Li, Wang, Wang, and
  Chen]{hu2021lora}
E.~J. Hu, Y.~Shen, P.~Wallis, Z.~Allen-Zhu, Y.~Li, S.~Wang, L.~Wang, and
  W.~Chen.
\newblock {LoRA}: Low-rank adaptation of large language models.
\newblock In \emph{ICLR}, 2022.

\bibitem[Hu(2024)]{hu2024animate}
L.~Hu.
\newblock {Animate Anyone}: Consistent and controllable image-to-video
  synthesis for character animation.
\newblock In \emph{CVPR}, 2024.

\bibitem[Huang et~al.(2022)Huang, Jin, Yi, and Sigal]{huang2022layered}
J.~Huang, Y.~Jin, K.~M. Yi, and L.~Sigal.
\newblock Layered controllable video generation.
\newblock In \emph{ECCV}, 2022.

\bibitem[Kingma(2015)]{kingma2014adam}
D.~P. Kingma.
\newblock Adam: A method for stochastic optimization.
\newblock In \emph{ICLR}, 2015.

\bibitem[Li et~al.(2024)Li, Wang, Zhang, Wang, Yuan, Xie, Zou, and
  Shan]{li2024imageConductor}
Y.~Li, X.~Wang, Z.~Zhang, Z.~Wang, Z.~Yuan, L.~Xie, Y.~Zou, and Y.~Shan.
\newblock {Image Conductor}: Precision control for interactive video synthesis.
\newblock \emph{arXiv preprint arXiv:2406.15339}, 2024.

\bibitem[Meng et~al.(2024)Meng, Liao, Tan, Shao, Lu, Zhang, Cheng, Li, Qiao,
  and Luo]{meng2024towards}
F.~Meng, J.~Liao, X.~Tan, W.~Shao, Q.~Lu, K.~Zhang, Y.~Cheng, D.~Li, Y.~Qiao,
  and P.~Luo.
\newblock Towards world simulator: Crafting physical commonsense-based
  benchmark for video generation.
\newblock \emph{arXiv preprint arXiv:2410.05363}, 2024.

\bibitem[Park et~al.(2024)Park, Jeong, Lee, and Ye]{park2024spectral}
G.~Y. Park, H.~Jeong, S.~W. Lee, and J.~C. Ye.
\newblock Spectral motion alignment for video motion transfer using diffusion
  models.
\newblock \emph{arXiv preprint arXiv:2403.15249}, 2024.

\bibitem[Polyak et~al.(2024)Polyak, Zohar, Brown, Tjandra, Sinha, Lee, Vyas,
  Shi, Ma, Chuang, et~al.]{polyak2024movie}
A.~Polyak, A.~Zohar, A.~Brown, A.~Tjandra, A.~Sinha, A.~Lee, A.~Vyas, B.~Shi,
  C.-Y. Ma, C.-Y. Chuang, et~al.
\newblock {Movie Gen}: A cast of media foundation models.
\newblock \emph{arXiv preprint arXiv:2410.13720}, 2024.

\bibitem[Radford et~al.(2021)Radford, Kim, Hallacy, Ramesh, Goh, Agarwal,
  Sastry, Askell, Mishkin, Clark, et~al.]{clip}
A.~Radford, J.~W. Kim, C.~Hallacy, A.~Ramesh, G.~Goh, S.~Agarwal, G.~Sastry,
  A.~Askell, P.~Mishkin, J.~Clark, et~al.
\newblock Learning transferable visual models from natural language
  supervision.
\newblock In \emph{ICML}, 2021.

\bibitem[Ramesh et~al.(2021)Ramesh, Pavlov, Goh, Gray, Voss, Radford, Chen, and
  Sutskever]{ramesh2021zero}
A.~Ramesh, M.~Pavlov, G.~Goh, S.~Gray, C.~Voss, A.~Radford, M.~Chen, and
  I.~Sutskever.
\newblock Zero-shot text-to-image generation.
\newblock In \emph{ICML}, 2021.

\bibitem[Ren et~al.(2021)Ren, Chai, Woodford, Olszewski, and
  Tulyakov]{ren2021flow}
J.~Ren, M.~Chai, O.~J. Woodford, K.~Olszewski, and S.~Tulyakov.
\newblock Flow guided transformable bottleneck networks for motion retargeting.
\newblock In \emph{CVPR}, 2021.

\bibitem[Saharia et~al.(2022)Saharia, Chan, Saxena, Li, Whang, Denton,
  Ghasemipour, Gontijo~Lopes, Karagol~Ayan, Salimans,
  et~al.]{saharia2022photorealistic}
C.~Saharia, W.~Chan, S.~Saxena, L.~Li, J.~Whang, E.~L. Denton, K.~Ghasemipour,
  R.~Gontijo~Lopes, B.~Karagol~Ayan, T.~Salimans, et~al.
\newblock Photorealistic text-to-image diffusion models with deep language
  understanding.
\newblock In \emph{NeurIPS}, 2022.

\bibitem[Salimans and Ho(2022)]{salimans2022progressive}
T.~Salimans and J.~Ho.
\newblock Progressive distillation for fast sampling of diffusion models.
\newblock In \emph{ICLR}, 2022.

\bibitem[Siarohin et~al.(2019)Siarohin, Lathuili{\`e}re, Tulyakov, Ricci, and
  Sebe]{siarohin2019first}
A.~Siarohin, S.~Lathuili{\`e}re, S.~Tulyakov, E.~Ricci, and N.~Sebe.
\newblock First order motion model for image animation.
\newblock In \emph{NeurIPS}, 2019.

\bibitem[Siarohin et~al.(2021)Siarohin, Woodford, Ren, Chai, and
  Tulyakov]{siarohin2021motion}
A.~Siarohin, O.~J. Woodford, J.~Ren, M.~Chai, and S.~Tulyakov.
\newblock Motion representations for articulated animation.
\newblock In \emph{CVPR}, 2021.

\bibitem[Singer et~al.(2023)Singer, Polyak, Hayes, Yin, An, Zhang, Hu, Yang,
  Ashual, Gafni, et~al.]{singer2022make}
U.~Singer, A.~Polyak, T.~Hayes, X.~Yin, J.~An, S.~Zhang, Q.~Hu, H.~Yang,
  O.~Ashual, O.~Gafni, et~al.
\newblock {Make-A-Video}: Text-to-video generation without text-video data.
\newblock In \emph{ICLR}, 2023.

\bibitem[Song et~al.(2019)Song, Zhu, Li, Wang, and Qi]{song2018talking}
Y.~Song, J.~Zhu, D.~Li, X.~Wang, and H.~Qi.
\newblock Talking face generation by conditional recurrent adversarial network.
\newblock In \emph{IJCAI}, 2019.

\bibitem[Song et~al.(2021)Song, Sohl-Dickstein, Kingma, Kumar, Ermon, and
  Poole]{song2020score}
Y.~Song, J.~Sohl-Dickstein, D.~P. Kingma, A.~Kumar, S.~Ermon, and B.~Poole.
\newblock Score-based generative modeling through stochastic differential
  equations.
\newblock In \emph{ICLR}, 2021.

\bibitem[Unterthiner et~al.(2018)Unterthiner, Van~Steenkiste, Kurach, Marinier,
  Michalski, and Gelly]{unterthiner2018fvd}
T.~Unterthiner, S.~Van~Steenkiste, K.~Kurach, R.~Marinier, M.~Michalski, and
  S.~Gelly.
\newblock Towards accurate generative models of video: A new metric \&
  challenges.
\newblock \emph{arXiv preprint arXiv:1812.01717}, 2018.

\bibitem[Valevski et~al.(2024)Valevski, Leviathan, Arar, and
  Fruchter]{valevski2024diffusion}
D.~Valevski, Y.~Leviathan, M.~Arar, and S.~Fruchter.
\newblock Diffusion models are real-time game engines.
\newblock \emph{arXiv preprint arXiv:2408.14837}, 2024.

\bibitem[Wang et~al.(2024{\natexlab{a}})Wang, Zhang, Zou, Zeng, Wei, Yuan, and
  Li]{wang2024boximator}
J.~Wang, Y.~Zhang, J.~Zou, Y.~Zeng, G.~Wei, L.~Yuan, and H.~Li.
\newblock Boximator: Generating rich and controllable motions for video
  synthesis.
\newblock \emph{arXiv preprint arXiv:2402.01566}, 2024{\natexlab{a}}.

\bibitem[Wang et~al.(2024{\natexlab{b}})Wang, Yuan, Zhang, Chen, Wang, Zhang,
  Shen, Zhao, and Zhou]{wang2024videocomposer}
X.~Wang, H.~Yuan, S.~Zhang, D.~Chen, J.~Wang, Y.~Zhang, Y.~Shen, D.~Zhao, and
  J.~Zhou.
\newblock {VideoComposer}: Compositional video synthesis with motion
  controllability.
\newblock In \emph{NeurIPS}, 2024{\natexlab{b}}.

\bibitem[Wang et~al.(2022)Wang, Li, Li, He, Huang, Zhao, Zhang, Xu, Liu, Wang,
  et~al.]{wang2022internvideo}
Y.~Wang, K.~Li, Y.~Li, Y.~He, B.~Huang, Z.~Zhao, H.~Zhang, J.~Xu, Y.~Liu,
  Z.~Wang, et~al.
\newblock {InternVideo}: General video foundation models via generative and
  discriminative learning.
\newblock \emph{arXiv preprint arXiv:2212.03191}, 2022.

\bibitem[Wang et~al.(2024{\natexlab{c}})Wang, Li, Li, Yu, He, Chen, Pei, Zheng,
  Xu, Wang, et~al.]{wang2024internvideo2}
Y.~Wang, K.~Li, X.~Li, J.~Yu, Y.~He, G.~Chen, B.~Pei, R.~Zheng, J.~Xu, Z.~Wang,
  et~al.
\newblock {InternVideo2}: Scaling video foundation models for multimodal video
  understanding.
\newblock In \emph{ECCV}, 2024{\natexlab{c}}.

\bibitem[Wang et~al.(2024{\natexlab{d}})Wang, Yuan, Wang, Li, Chen, Xia, Luo,
  and Shan]{wang2024motionctrl}
Z.~Wang, Z.~Yuan, X.~Wang, Y.~Li, T.~Chen, M.~Xia, P.~Luo, and Y.~Shan.
\newblock {MotionCtrl}: A unified and flexible motion controller for video
  generation.
\newblock In \emph{ACM SIGGRAPH}, 2024{\natexlab{d}}.

\bibitem[Wu et~al.(2024)Wu, Li, Gu, Zhao, He, Zhang, Shou, Li, Gao, and
  Zhang]{wu2025draganything}
W.~Wu, Z.~Li, Y.~Gu, R.~Zhao, Y.~He, D.~J. Zhang, M.~Z. Shou, Y.~Li, T.~Gao,
  and D.~Zhang.
\newblock {DragAnything}: Motion control for anything using entity
  representation.
\newblock In \emph{ECCV}, 2024.

\bibitem[Xiang et~al.(2024)Xiang, Liu, Gu, Gao, Ning, Zha, Feng, Tao, Hao, Shi,
  et~al.]{xiang2024pandora}
J.~Xiang, G.~Liu, Y.~Gu, Q.~Gao, Y.~Ning, Y.~Zha, Z.~Feng, T.~Tao, S.~Hao,
  Y.~Shi, et~al.
\newblock Pandora: Towards general world model with natural language actions
  and video states.
\newblock \emph{arXiv preprint arXiv:2406.09455}, 2024.

\bibitem[Xiao et~al.(2024)Xiao, Zhou, Yang, and Pan]{xiao2024video}
Z.~Xiao, Y.~Zhou, S.~Yang, and X.~Pan.
\newblock Video diffusion models are training-free motion interpreter and
  controller.
\newblock In \emph{NeurIPS}, 2024.

\bibitem[Yang et~al.(2024)Yang, Du, Ghasemipour, Tompson, Schuurmans, and
  Abbeel]{yang2023learning}
M.~Yang, Y.~Du, K.~Ghasemipour, J.~Tompson, D.~Schuurmans, and P.~Abbeel.
\newblock Learning interactive real-world simulators.
\newblock In \emph{ICLR}, 2024.

\bibitem[Yuan et~al.(2024)Yuan, Gundavarapu, Zhao, Zhou, Cui, Jiang, Yang, Jia,
  Weyand, Friedman, et~al.]{yuan2023videoglue}
L.~Yuan, N.~B. Gundavarapu, L.~Zhao, H.~Zhou, Y.~Cui, L.~Jiang, X.~Yang,
  M.~Jia, T.~Weyand, L.~Friedman, et~al.
\newblock {VideoGLUE}: Video general understanding evaluation of foundation
  models.
\newblock \emph{TMLR}, 2024.

\bibitem[Zhao et~al.(2018)Zhao, Peng, Tian, Kapadia, and
  Metaxas]{zhao2018learning}
L.~Zhao, X.~Peng, Y.~Tian, M.~Kapadia, and D.~Metaxas.
\newblock Learning to forecast and refine residual motion for image-to-video
  generation.
\newblock In \emph{ECCV}, 2018.

\bibitem[Zhao et~al.(2024{\natexlab{a}})Zhao, Gundavarapu, Yuan, Zhou, Yan,
  Sun, Friedman, Qian, Weyand, Zhao, et~al.]{zhao2024videoprism}
L.~Zhao, N.~B. Gundavarapu, L.~Yuan, H.~Zhou, S.~Yan, J.~J. Sun, L.~Friedman,
  R.~Qian, T.~Weyand, Y.~Zhao, et~al.
\newblock {VideoPrism}: A foundational visual encoder for video understanding.
\newblock In \emph{ICML}, 2024{\natexlab{a}}.

\bibitem[Zhao et~al.(2024{\natexlab{b}})Zhao, Gu, Wu, Zhang, Liu, Wu, Keppo,
  and Shou]{zhao2023motiondirector}
R.~Zhao, Y.~Gu, J.~Z. Wu, D.~J. Zhang, J.-W. Liu, W.~Wu, J.~Keppo, and M.~Z.
  Shou.
\newblock {MotionDirector}: Motion customization of text-to-video diffusion
  models.
\newblock In \emph{ECCV}, 2024{\natexlab{b}}.

\end{thebibliography}

\appendix

\newpage
\onecolumn
\section*{Appendix}
\section{Author Contributions}
We list authors alphabetically by last name. Please direct all correspondence to Ting Liu (\url{liuti@google.com}) and Long Zhao (\url{longzh@google.com}).

\subsection*{Core Contributors}
\begin{itemize}
\item \textbf{Ting Liu:} \vcbd (VCBD) concept, project leadership, dataset curation, evaluation, action model research, model demo.
\item \textbf{Jennifer J. Sun:} Generation model research, infrastructure, evaluation.
\item \textbf{Yihong Sun:} Generation model research, action model research, infrastructure, evaluation, model demo.
\item \textbf{Liangzhe Yuan:} VCBD concept, action model research, evaluation.
\item \textbf{Long Zhao:} Project leadership, generation model research, action model research, evaluation.
\item \textbf{Hao Zhou:} Action model research, evaluation, infrastructure.
\end{itemize}

\subsection*{Partial Contributors and Advisors}
\begin{itemize}
\item \textbf{Bharath Hariharan:} Technical advice.
\item \textbf{Xuhui Jia:} Technical advice.
\item \textbf{Yandong Li:} Technical advice.
\end{itemize}

\subsection*{Sponsors}
\begin{itemize}
\item \textbf{Hartwig Adam:} Strategic advice.
\end{itemize}

\section{Improving Generation Quality}
\label{sec:improve_quality}

To support demonstration video with length longer than the generation length $T=16$, we follow the auto-regressive generation setup in WALT~\citep{walt}.
During model $\mG$ training, we swap out the input context image with a probability of $0.5$ and replace it with multiple consecutive frames to encourage a smoother continual generation.
At inference time, we cut long demonstration video into multiple segments as needed, each of length $T$.
This allows the generated segment to remain aligned with the appropriate demonstration segment.
Here, the first segment is generated from the input context image, while the subsequent segments are conditioned on the last $4$ generated frames from a previous segment.

During training, we enable self-conditioning~\citep{chen2022analog} with a probability of $0.9$ and randomly mask out the control signal with a probability of $0.2$.
At inference time, we adopt classifier-free guidance consistent with \citet{saharia2022photorealistic} with a guidance weight of $1.25$ and drop both the self-conditioning and control signal for the unconditional generation.

\begin{table*}[htb]
\caption{Human evaluation instructions and rubrics.}
\centering
\begin{tabular}{ll}
\toprule
Instructions & \\
\midrule
\multicolumn{2}{l}{\parbox{0.95\linewidth}{\emph{Given an initial image (context frame) and a demonstration video (reference video), the task of Video Creation by Demonstration is defined as to create a plausible video clip initiating from the context frame and contains similar content dynamics as in the demonstration video.}}} \\
& \\
\multicolumn{2}{l}{\parbox{0.95\linewidth}{\emph{In this user study, you will be provided a reference video, a context frame, and two generated videos from two methods side-by-side in each turn. You would assess the generated video quality on the following three rubrics and determine which one better recreates the actions and dynamics of a given demonstration video, while also appearing realistic and continuous from a context image.}}} \\
& \\
\midrule
Rubrics & Descriptions\\
\midrule
Visual Quality & \parbox{0.7\linewidth}{\emph{How realistic does the video look compared to a real-world video? Consider factors like smooth motion, accurate details, and believable physics. Choose the video that is better.}}\\
& \\
Action Transferability & \parbox{0.7\linewidth}{\emph{How well does the generated video recreate the actions and movements shown in the reference video? Note that the action concept can include the camera motions. Choose the video that is better.}}\\
& \\
Frame Continuity & \parbox{0.7\linewidth}{\emph{How seamlessly does the generated video flow from the provided context frame? Does it look like a natural continuation of the scene? Choose the video that is better.}}\\
& \\
\bottomrule
\end{tabular}\label{tab:user}
\end{table*} 
\section{Human Evaluation Setup}
\label{sec:user_study}

\subsection{Prompt Selection}
For all three datasets, we consider the demonstration video to be between $16$ to $24$ frames at $12$ frames per second (FPS) (Something-Something v2 and Epic Kitchens 100) or $10$ FPS (Fractal).
For Something-Something v2~\citep{goyal2017something}, we select the top-10 action classes.
For each action class, we manually select $20$ out of $500$ randomly sampled pairs for human evaluation.
For Epic Kitchens 100~\citep{ek100}, we apply automatic filtering to first narrow down the search. For each video, we retrieve top-$1$ video with the same action label and different participant ID via first-frame CLIP~\citep{clip} similarity. Then, $20$ pairs are randomly sampled from top-$25$ action classes. Finally, $20$ examples are manually selected from a pool of pairs with CLIP similarity greater than $0.9$.
For Fractal, we sample $500$ same-action pairs uniformly from each of the $8$ action classes (defined by the verb and preposition from the associated captions) and conduct manual selection in random order to select $20$ pairs for evaluation.
The main manual selection principle is to confirm that the action concept in the demonstration video can be potentially applied in a reasonable way from the context image. This process is conducted without any considerations of the methods to be evaluated.

From each selected video pair, one video is randomly selected as demonstration, and the first of frame of the other video is used as the context image to construct a prompt.

\subsection{Rating Instructions and Rubrics}
In Table~\ref{tab:user}, we list the instructions and rubrics shown to the human raters before presenting them with the generated videos for side-by-side comparisons.

\subsection{Additional Details}
When presenting the visual examples, we show \textit{\textbf{Demonstration Video}}, \textit{\textbf{Context Image}}, \textit{\textbf{Video A}}, and \textit{\textbf{Video B}} in order.
As we conduct the user study via Colab, we randomly assign \textit{Video A} to be either \ours or a baseline method. We hired $5$ human raters to examine $2$ baselines to compare against \ours on $3$ datasets, $20$ examples per dataset, and $3$ metrics per example.
In total, we collected $1800$ human preferences.

\section{Details on Baselines}
\paragraph{MotionDirector~\citep{zhao2023motiondirector}.}
When testing MotionDirector on Something-Something v2 (SSv2)~\citep{goyal2017something}, Epic Kitchens 100~\citep{ek100}, and Fractal~\citep{fractal}, we use the official code published by the authors at \url{https://github.com/showlab/MotionDirector}.
We first train the spatial path with the context image for $300$ epochs with input resolution being $384\times384$.
Then we train the temporal path following the $16$-frame single video setup.
During inference, a noise prior of $0.0$ is applied.
Ground truth caption labels are used as the prompt for both training and inference.

\paragraph{WALT~\citep{walt}.}
We fine-tune WALT on each dataset in accordance with the official procedures and use ground truth captions to tune video generation from context image and text caption.
We keep all hyper-parameters the same and fine-tune for $200$k steps.
During inference, the ground truth caption of the demonstration video is used along with the context image for video generation.
As shown in the visualizations, WALT generations, while preserving context consistency, carry out ungenuine action due to limited expressiveness in text descriptions.

\section{Qualitative Results}
\label{sec:vis}
The original videos for creating Figures~\ref{fig:teaser} and \ref{fig:video_collage} can be found at \url{\projectpage}. The original videos for creating Figures \ref{fig:ablation} and \ref{fig:qual} can be found at \url{\projectpage/additional.html}, along with additional examples.

\section{Controllability by Demonstration}
\label{sec:alt_future}
\begin{figure*}[t]
    \centering
    \includegraphics[width=\linewidth]{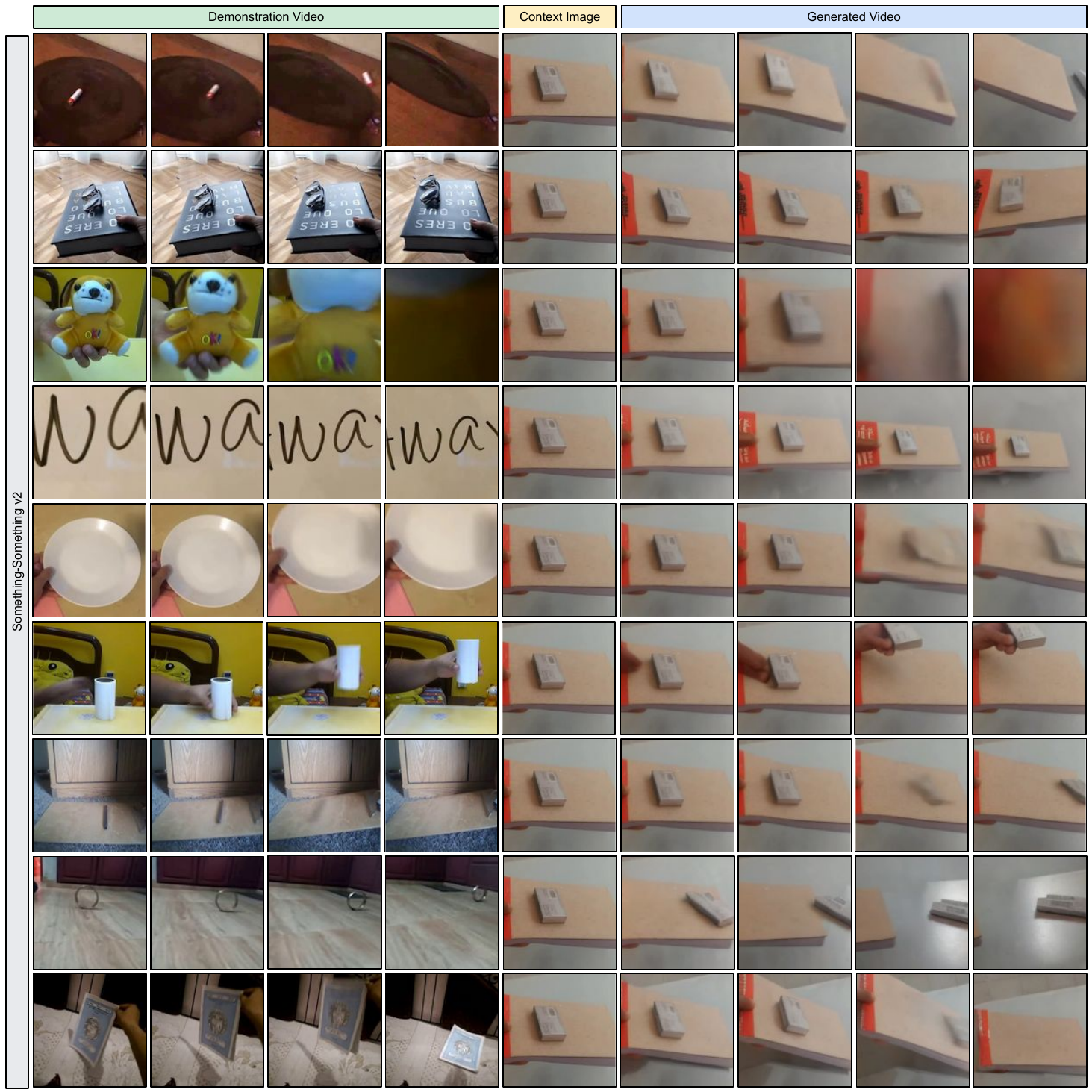}
    \caption{Qualitative results of driving alternative generation from the same context image with different demonstration videos from the Something-Something v2 dataset~\citep{goyal2017something}. 
}
    \label{fig:alt_future_ss1}
\end{figure*}

\begin{figure*}[t]
    \centering
    \includegraphics[width=\linewidth]{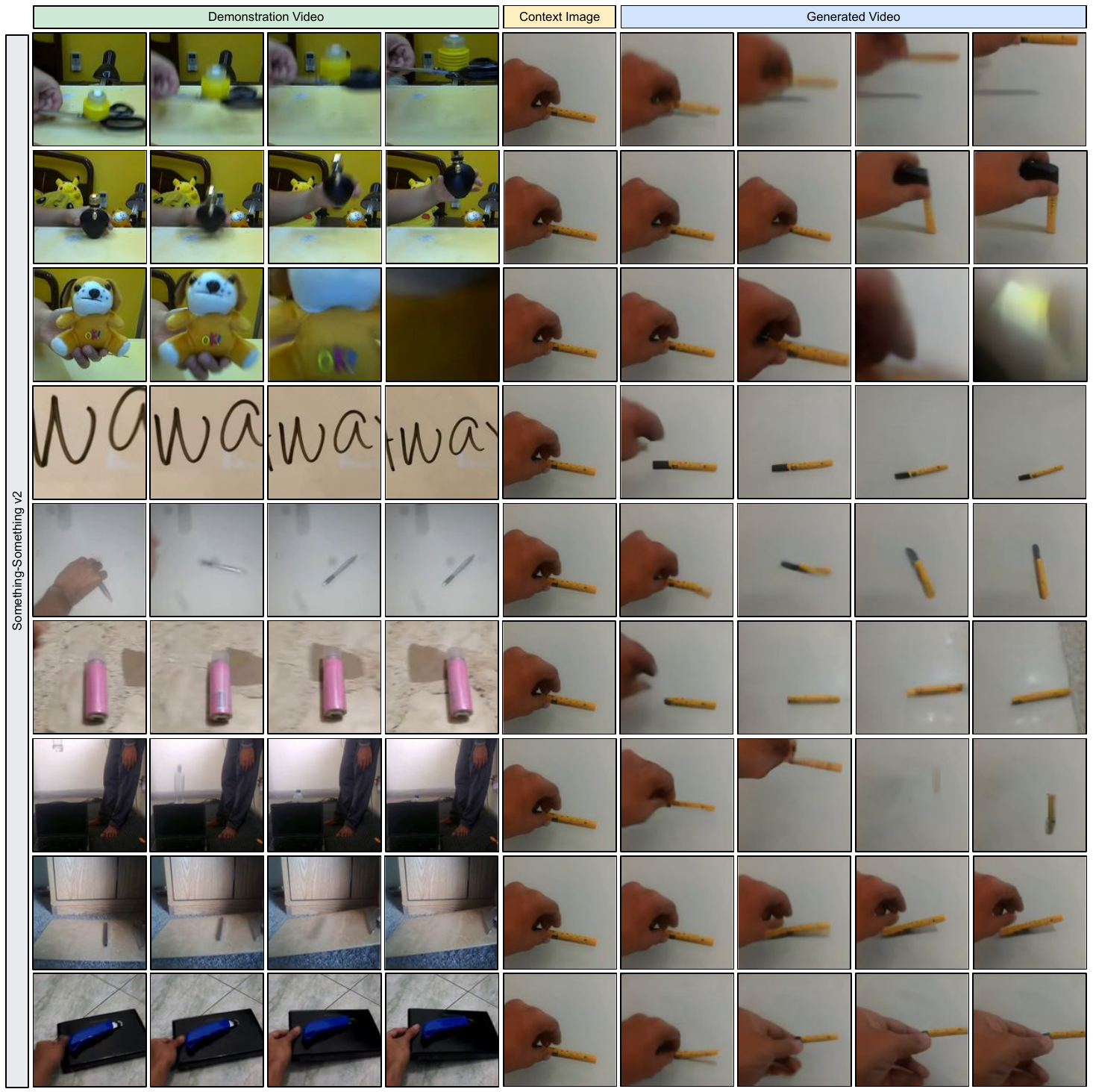}
    \caption{Qualitative results of driving alternative generation from the same context image with different demonstration videos from the Something-Something v2 dataset~\citep{goyal2017something}. 
}
    \label{fig:alt_future_ss2}
\end{figure*}

\begin{figure*}[t]
    \centering
    \includegraphics[width=\linewidth]{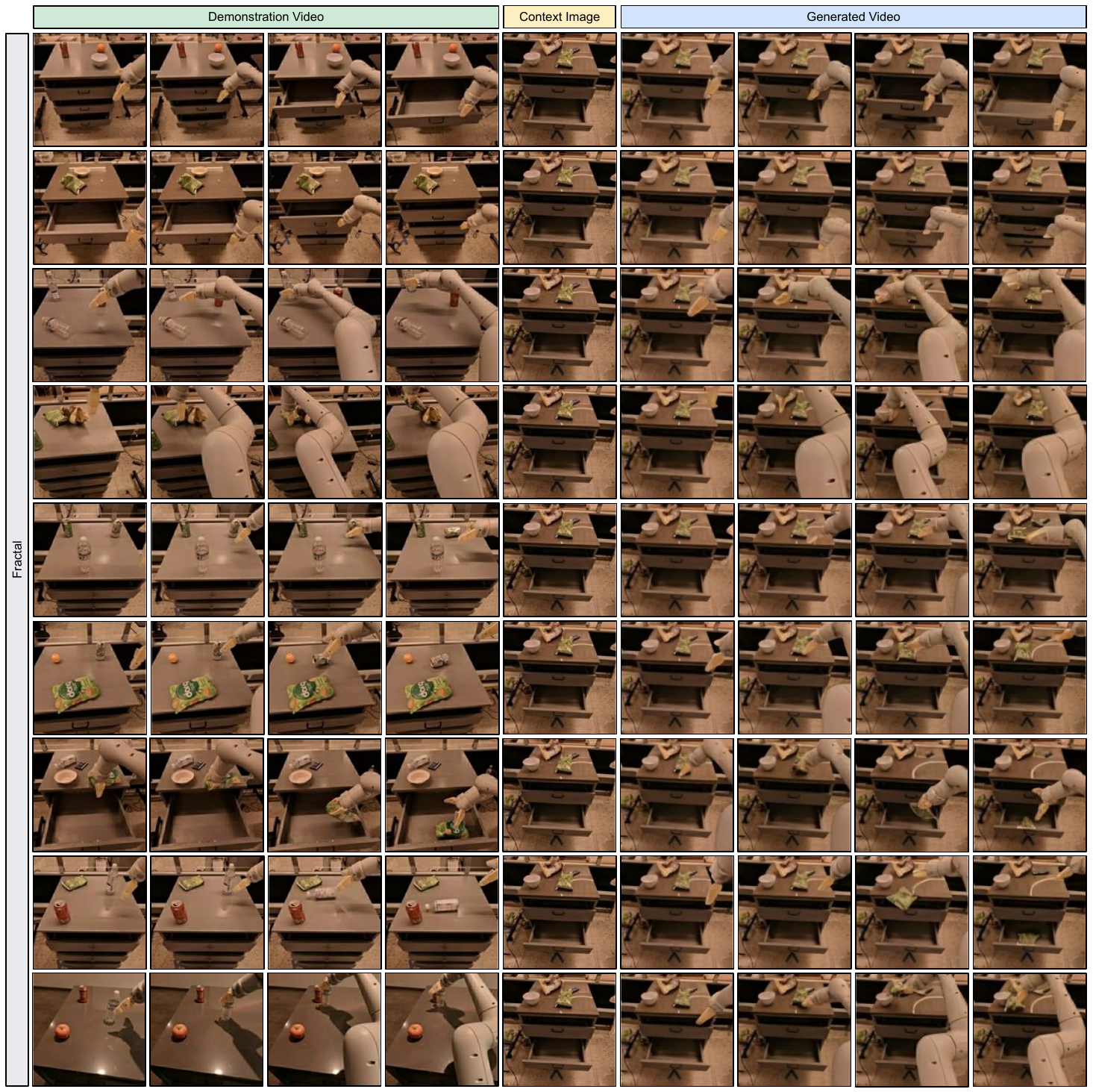}
    \caption{Qualitative results of driving alternative generation from the same context image with different demonstration videos from the Fractal dataset~\citep{fractal}.
}
    \label{fig:alt_future_fr1}
\end{figure*}

\begin{figure*}[t]
    \centering
    \includegraphics[width=\linewidth]{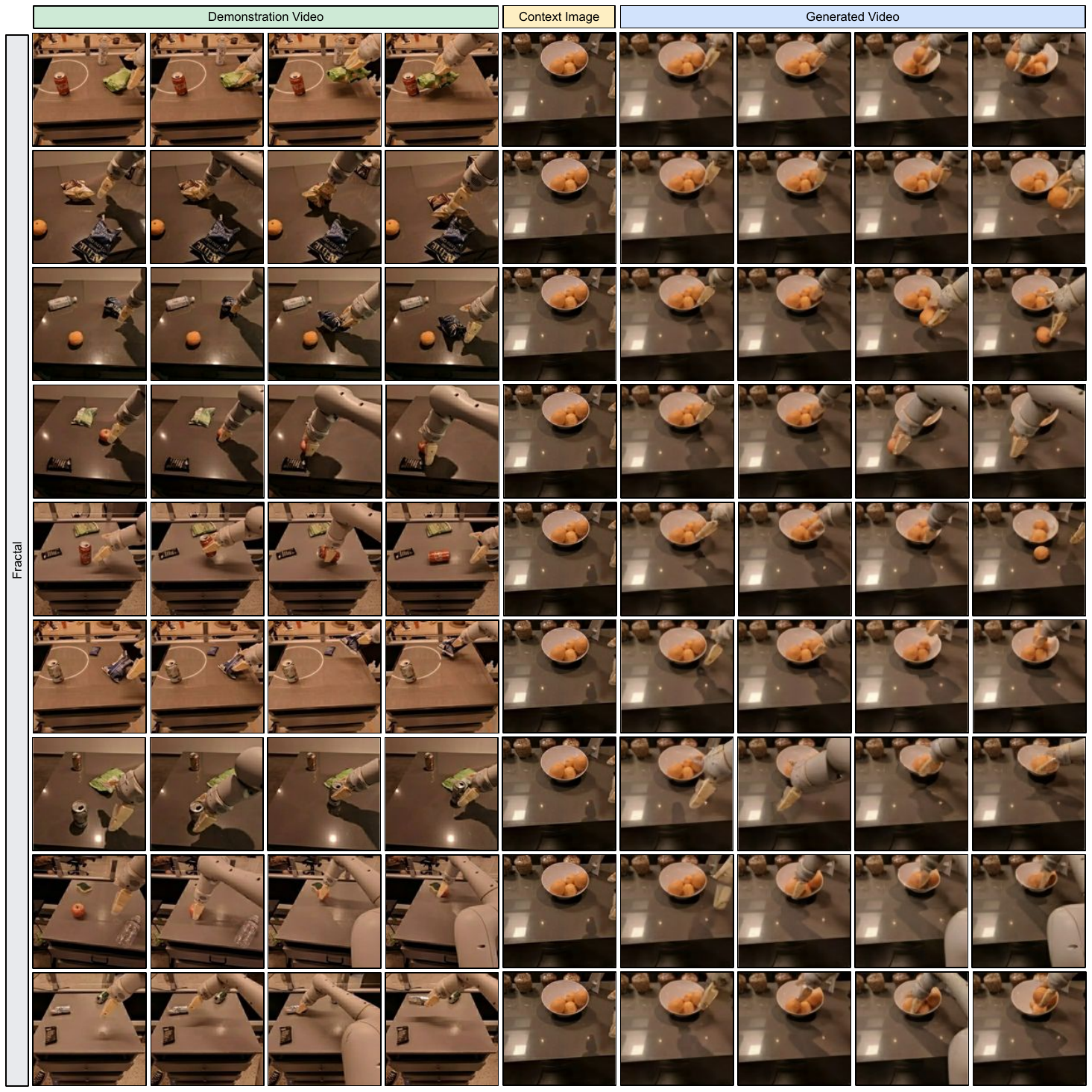}
    \caption{Qualitative results of driving alternative generation from the same context image with different demonstration videos from the Fractal dataset~\citep{fractal}.
}
    \label{fig:alt_future_fr2}
\end{figure*}

As shown in Figures~\ref{fig:alt_future_ss1}, \ref{fig:alt_future_ss2}, \ref{fig:alt_future_fr1}, and \ref{fig:alt_future_fr2}, we showcase the controllability of \ours using different demonstration videos for the same context image. The video samples can be found at \url{\projectpage/\#cbd}.

\section{Failure Cases}
\label{sec:limitation}
\begin{figure*}[t]
    \centering
    \includegraphics[width=\linewidth]{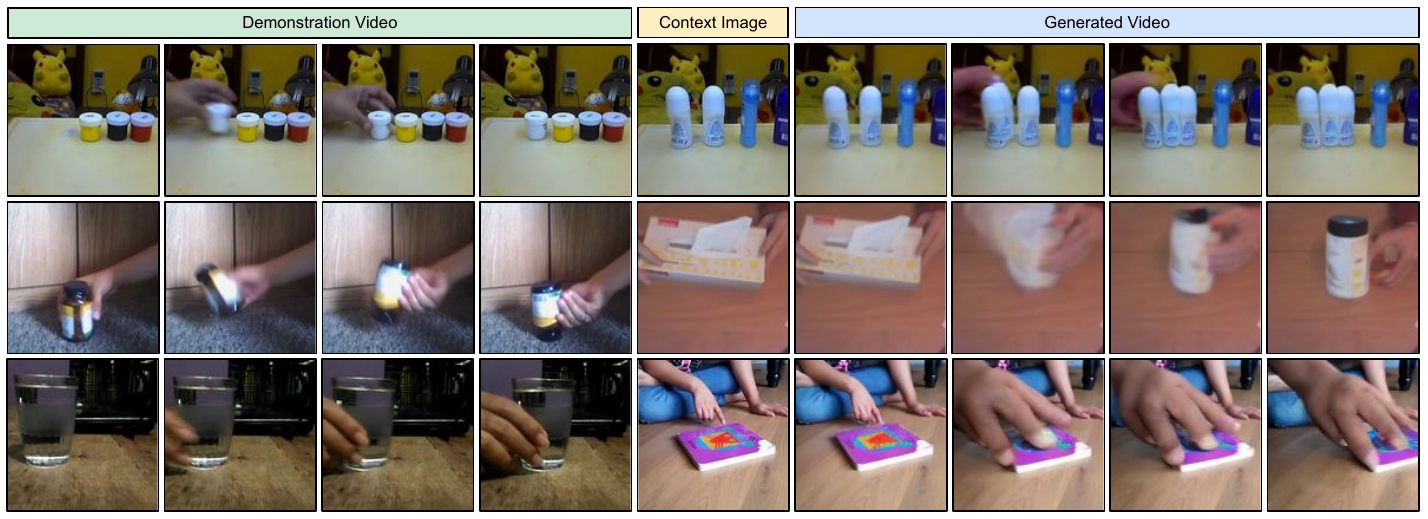}
    \caption{Failure cases generated by \ours. 
}
    \label{fig:failure}
\end{figure*}

As shown in Figure~\ref{fig:failure}, we identify three primary failure modes of our method. In the first row, we show a case where the semantics of the action concept in the demonstration videos are not fully carried out. Specifically, the generated object is placed ``in-between'' the existing objects instead of ``next-to'' them. In the second row, we show a case where permanence is not held when the object in the demonstration video undergoes fast appearance changes. Here, fast object rotation causes appearance leakage in the generation. In the third row, we show inconsistent generations where the demonstration videos and context images are mis-matched significantly. On the left, the perspectives of the moving hand are mis-matched and cause another hand in the same demonstrated perspective to be generated.

\end{document}